\documentclass[10pt,twocolumn,letterpaper]{article}

\usepackage{cvpr}              
\usepackage[ruled,vlined]{algorithm2e}
\usepackage{float}
\usepackage{multirow}
\usepackage{adjustbox} 
\usepackage{booktabs}
\usepackage{tabularx}
\usepackage{cuted}
\usepackage{graphicx}
\usepackage{subcaption}  
\usepackage{pifont}   
\usepackage{placeins}
\usepackage[accsupp]{axessibility}

\definecolor{cvprblue}{rgb}{0.21,0.49,0.74}
\usepackage[pagebackref,breaklinks,colorlinks,allcolors=cvprblue]{hyperref}

\def\paperID{45918} 
\def\confName{CVPR}
\def\confYear{2026}

\title{
AREA3D: Active Reconstruction Agent with Unified Feed-Forward 3D Perception and Vision-Language Guidance
}

\author{
Tianling Xu$^{1,2}$\quad
Shengzhe Gan$^{1}$ \quad
Leslie Gu$^{2}$ \quad
Yuelei Li$^{3}$ \quad
Fangneng Zhan$^{4}$ \quad
Hanspeter Pfister$^{2}$\\[0.3em]
$^{1}$Southern University of Science and Technology\\
$^{2}$Harvard University\\
$^{3}$California Institute of Technology\\
$^{4}$Massachusetts Institute of Technology
}

\begin{document}

\maketitle


\begin{strip}
  \centering
  \includegraphics[width=\textwidth]{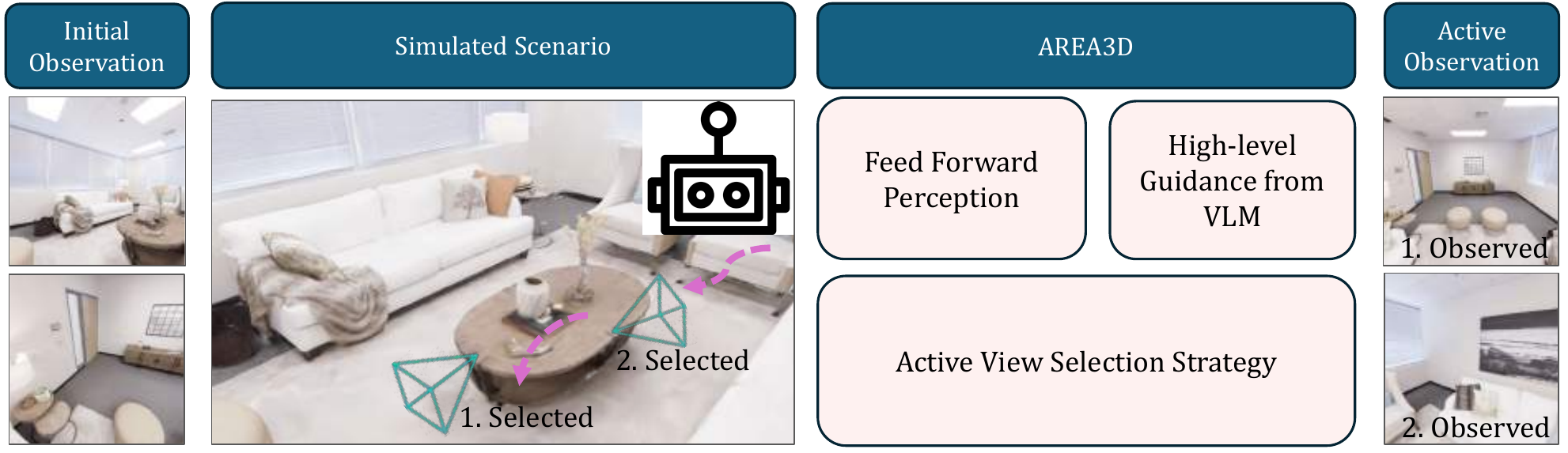}
  \captionof{figure}{
    \textbf{Overview of our approach.}
    We propose AREA3D, an active reconstruction agent, which unifies two complementary signals of feed-forward 3D perception and vision-language guidance to decide the
    next best views under tight view budgets.
    AREA3D efficiently reconstructs high-fidelity geometry from sparse observations by actively choosing the most informative viewpoints.
  }
  \label{fig:teaser}
\end{strip}

\begin{abstract}
Active 3D reconstruction enables an agent to autonomously select viewpoints to build accurate and complete scene geometry, rather than passively reconstructing from pre-collected datasets. Existing active reconstruction methods often rely on geometric heuristics, which may lead to redundant observations without improving reconstruction quality. To address this limitation, we propose \textbf{AREA3D}, an active 3D reconstruction agent that leverages feed-forward 3D models and data-driven vision-language guidance.The framework decouples view uncertainty modeling from the feed-forward 3D model and the pretrained Vision-Language model, enabling precise uncertainty estimation without costly online optimization. Moreover, the integrated Vision-Language Model provides high-level semantic guidance for exploration beyond purely geometric cues. Extensive experiments on both scene-level and object-level benchmarks (Replica and OmniObject3D) demonstrate that AREA3D achieves state-of-the-art reconstruction accuracy, especially in sparse views.

\end{abstract}

\section{Introduction}
\label{sec:intro}
Active and autonomous exploration constitutes a cornerstone of intelligent embodied agents, enabling them to perceive, interpret, and interact with complex environments. Within this broader context, active reconstruction represents a critical subtask that demands the integration of 3D perception, viewpoint selection, and embodied execution.

Traditional active reconstruction methods~\cite{lee2022uncertainty,fisherrf,activeNBV,goli2024bayes} rely on handcrafted criteria, such as surface coverage, voxel occupancy, or view overlap heuristics, to guide view selection. However, these handcrafted metrics provide no direct awareness of under-reconstructed or missing regions. 

As a result, they often overestimate areas that are already well-covered but poorly reconstructed, while failing to identify holes or unseen surfaces caused by occlusion or limited viewpoints. This mismatch leads to redundant observations and incomplete scene geometry. To overcome this, recent work~\cite{air,activegs,activesplat,naruto,fisherrf} uses high-fidelity neural representations, such as 3D Gaussian Splatting \cite{3dgs} and Neural Radiance Field \cite{nerf}, to model uncertainty or information gain. However, without robust data-driven priors, their performance degrades under sparse observations, while the computational cost becomes prohibitive when dense observations are available. Consequently, current approaches must carefully balance reconstruction quality and efficiency, limiting their scalability and flexibility. 

We propose \textbf{AREA3D}, an active reconstruction agent, which unifies two complementary signals of \textit{feed-forward 3D perception} and \textit{vision-language guidance} to decide the next best views under tight view budgets.
As shown in Figure~\ref{fig:teaser}, our framework features a dual-field uncertainty modeling mechanism. Unlike per-scene optimized NeRF or 3DGS, a data-driven feed-forward 3D model can directly provide geometric confidence to build a 3D geometric field about what is already well perceived, for both sparse views and dense views.
On the other hand, a vision–language model reasons about what is likely missing, highlighting occluded or unseen content. We combine these cues into a single, visibility-aware “where-to-look” field that guides the agent to a small number of high-value views, progressively closing coverage gaps. This design is lightweight, budget-aware, and can handle both object-centric tabletop scenes and room-scale scenes, yielding precise geometry with active exploration.

In general, we conclude our contributions as follows: 
\begin{itemize}
\item We propose AREA3D, an agent for active 3D reconstruction that unifies feed-forward 3D modeling and vision–language reasoning into a dual-field framework for uncertainty-aware perception and view planning. 
\item We construct a unified active reconstruction benchmark covering both object-centric and scene-level regimes, enabling consistent evaluation across scales and demonstrating strong robustness and generalization.
\item Extensive experiments demonstrate state-of-the-art accuracy and efficiency under tight budgets, with ablations confirming the complementary value of feed-forward 3D perception and high-level guidance from Vision Language Model.
\end{itemize}


\section{Related work}
\label{sec:formatting}
\subsection{Active Reconstruction}

Active reconstruction is a critical problem that lies at the intersection of active perception, 3D vision, and robotics. Traditional methods often formulate it as an active view planning task, where the agent selects informative viewpoints to minimize reconstruction uncertainty or maximize surface coverage. Table~\ref{tab:related_work_comparison} provides an overview of the uncertainty proxies employed by existing active reconstruction methods.


\noindent\textbf{Early Active View Selection.}~
Early approaches select observation viewpoints based on handcrafted geometric heuristics, such as voxel occupancy\cite{voxel}, view overlap, or frontier-based exploration VLFM \cite{vlfm,activemapping, NBV-Vol,volumetric-NBV}
. While effective in simple scenes, these metrics often rely on known geometry or occupancy grids and cannot capture complex scene ambiguity or perceptual uncertainty, frequently leading to redundant observations.

\noindent\textbf{Learning-Based \& Neural-Field–Driven Methods.}~
More recent learning-based methods employ policy networks or reinforcement learning \cite{gleam,gennbv,activeNBV,activenerf} to predict the next best view (NBV) from image features or partial reconstructions. A significant body of work has focused on leveraging the internal state of neural radiance fields (NeRF) \cite{nerf} to guide exploration. For example, some methods utilize uncertainty derived directly from the NeRF model's density field or rendering variance to optimize view trajectories \cite{lee2022uncertainty, naruto,neu-NBV,activenerf}. Beyond neural radiance fields, 3D Gaussian representations are increasingly explored for active reconstruction. Representative methods include ~\cite{activegs,activesplat,air}. These methods are inherently coupled to the online optimization, requiring costly gradient-based updates to assess information gain.

\noindent\textbf{Information-Theoretic Approaches.}~
In parallel to such uncertainty-driven and RL-based strategies, information-theoretic approaches~\cite{fisherrf,3dgs,nextbestpath} aim to reason about view utility in a more principled manner. FisherRF~\cite{fisherrf} exemplifies this direction by estimating the Fisher Information Matrix of radiance-field parameters ~\cite{3dgs} and selecting views via Expected Information Gain (EIG). While principled and ground-truth–free, it remains fundamentally tied to a high-density differentiable representation and inherits similar computational burdens, especially under sparse-view conditions where the field has not yet converged.
\subsection{Feed-forward 3D Reconstruction Model}
Feed-forward 3D reconstruction models aim to infer 3D structure from images in a single forward pass, avoiding costly iterative optimization. Unlike traditional multi-view stereo or differentiable implicit methods \cite{nerf}, these models provide fast reconstruction and can generalize across different objects or scenes. Representative approaches include \cite{dust3r}, \cite{mast3r}, \cite{cut3r}, and \cite{flare}, which use convolutional or transformer-based architectures to predict depth, point clouds, or meshes from single or multiple images.
The key advantage of these models for active perception is their ability to provide not only a 3D prediction but also a \textit{fast, associated uncertainty estimate} for that prediction. This uncertainty, derived from the model's internal confidence, can be computed in a single forward pass without requiring a partially converged 3D model or gradient-based optimization. We adopt VGGT\cite{vggt} as our backbone feed-forward model. It leverages priors learned from large-scale data and provides robust uncertainty predictions, enabling an efficient and scalable active reconstruction pipeline.


\begin{table}[H]
    \centering
    \captionsetup{justification=centering, singlelinecheck=false}
    \caption{Comparison of Different Uncertainty Proxies for Active Reconstruction.}
    \label{tab:related_work_comparison}
    
    \begin{tabularx}{\linewidth}{@{} >{\raggedright\arraybackslash\hsize=0.33\hsize}X >{\raggedright\arraybackslash\hsize=0.67\hsize}X @{}}
        \toprule
        \textbf{Uncertainty Proxy} & \textbf{Key Disadvantages} \\
        \midrule
        
        Geometric Heuristics \cite{gleam,vlfm,activemapping,volumetric-NBV,NBV-Vol}
         & 
        Ignores all perceptual and scene uncertainty; inefficient sampling with no guarantee of coverage. \\
        
        \addlinespace
        
        NeRF Rendering Variance
        \cite{activenerf,naruto,lee2022uncertainty,neu-NBV} & 
        Requires costly backpropagation through volumetric rendering; impractical for real-time planning; converges slowly from sparse views. \\
        
        \addlinespace
        
        3D Gaussian Splatting
        \cite{fisherrf,air,activegs,activesplat}, &
        Degrades severely under sparse-view inputs; coupled to online optimization to assess information gain. \\
        
        \addlinespace
        
        VLM Semantic Reasoning
        \cite{air,explorevlm,mllm,apvlm,imaginenav} &
        Lacks fine-grained geometry;  \space reasoning is non-metric; planning is slow and costly. \\
        \bottomrule
    \end{tabularx}
\end{table}
\subsection{Vision-Language Models for Robotics Planning}
Large Vision-Language Models (VLMs) have rapidly advanced in recent years, driven by the emergence of general-purpose multimodal foundation models ~\cite{internvl3,internvl35,qwen25,qwen_tech,move}. Building on this progress, recent works have explored using large VLMs and multi-modal models to support high-level task planning in robotics \cite{driess2023palm, zitkovich2023rt}. These models can take visual observations and language instructions as input to generate goal-directed actions, enabling flexible decision-making and generalization. VLM-based approaches have been applied to navigation, complex manipulation \cite{air, voxposer}, and multi-step planning, often serving as a high-level policy that guides downstream low-level controllers.
Beyond task planning, VLMs are increasingly explored for \textit{embodied perception and exploration}. Approaches such as \cite{apvlm,explorevlm,historyvlm,imaginenav,mllm} investigate how language-grounded reasoning can inform exploration strategies, scene understanding, or active perception. These methods highlight the potential of VLMs to capture semantic priors that are difficult to encode using purely geometric criteria.
In the context of active 3D reconstruction, VLMs offer complementary semantic reasoning to geometry-driven NBV planners. For example, AIR-Embodied~\cite{air} leverages VLMs to infer occlusions or explore semantically meaningful regions. Such high-level reasoning provides orthogonal guidance to purely geometric or uncertainty-based planners. In our experiments, we include a VLM-based planner as a baseline to examine this trade-off.

\section{Method}
\subsection{Overview}
\noindent\textbf{Problem Definition.}~
We consider \textbf{active 3D reconstruction} with a strict view budget.
Let $T$ be the total budget, and let $\mathcal{O}_0=\{(I_0,p_0),\dots\}$ denote the initial observations available to the agent, 
where each observation consists of an image $I_v$, its pose $p_v$, and optional depth $D_v$.
The remaining view budget is thus $T-|\mathcal{O}_0|$.

Selecting an additional view set $\mathcal{S}$ induces the posed observations
\[
\mathcal{O}(\mathcal{S}) \;=\; \mathcal{O}_0 \cup \{\, (I_v,\, p_v) \,\}_{v\in\mathcal{S}}.
\]
A reconstructor $R$ maps $\mathcal{O}(\mathcal{S})$ to a scene estimate 
\[
\hat{\mathcal{G}}(\mathcal{S}) = R\big(\mathcal{O}(\mathcal{S})\big).
\]
Given a task metric $\mathcal{Q}$ against the ground-truth scene $\mathcal{G}$, 
the budgeted view selection problem is formulated as
\[
\mathcal{S}^{\star} \;\in\; \arg\max_{\,|\mathcal{S}|\le T-|\mathcal{O}_0|}
\;\mathcal{Q}\!\left(\hat{\mathcal{G}}(\mathcal{S}),\, \mathcal{G}\right).
\]

In practice we solve this budgeted set selection by optimizing a principled surrogate objective that scores candidate views for their expected reconstruction benefit; the surrogate and scoring are detailed in the subsequent sections.

\vspace{0.5em}

\begin{figure*}[!t]
  \centering
  \includegraphics[width=\linewidth]{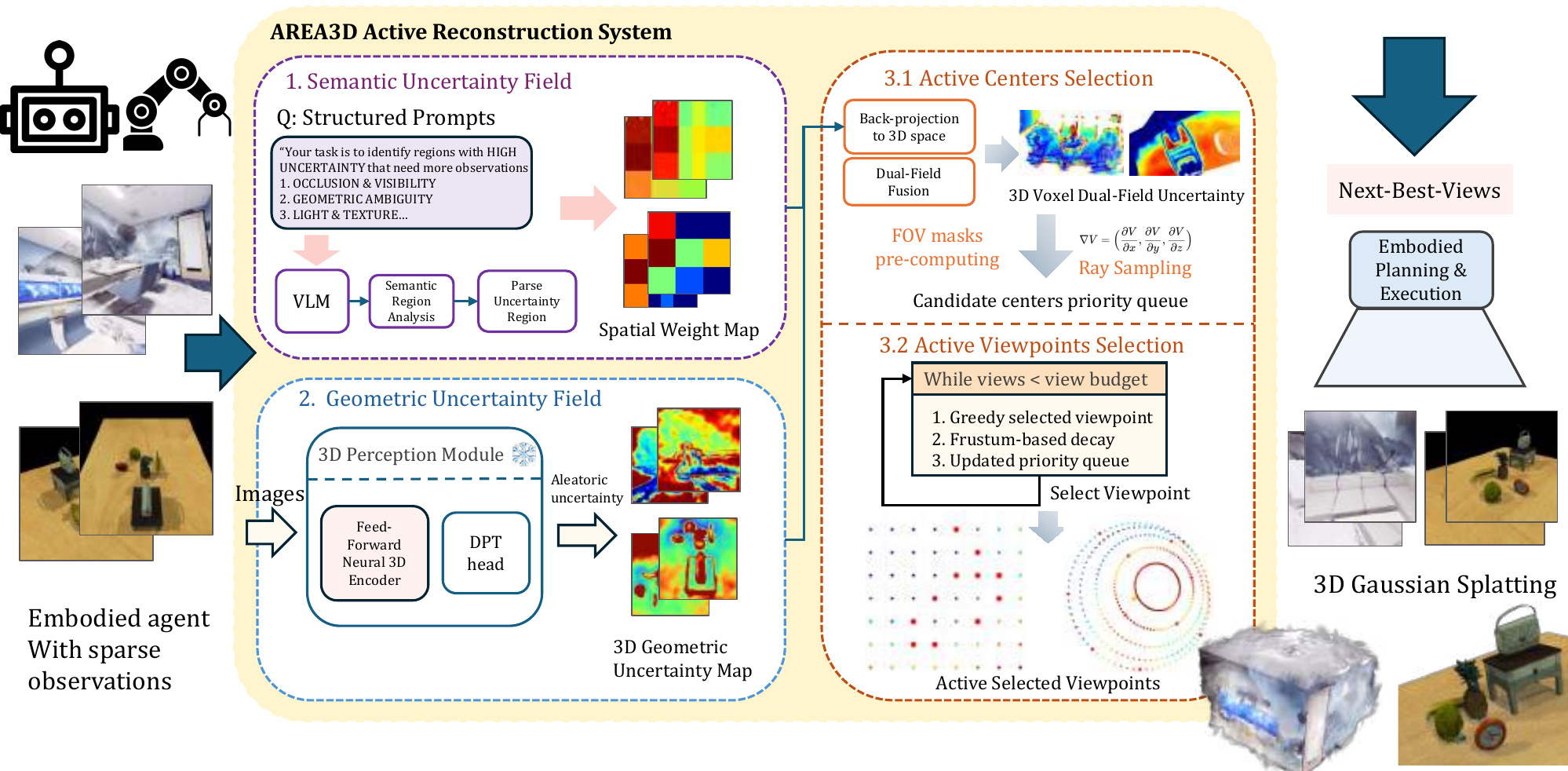}
  \vspace{-23pt} 
  \caption{Overview of the AREA3D pipeline. The framework integrates feed-forward 3D perception and vision-language guidance to actively select informative viewpoints and to reconstruct high-fidelity geometry via Gaussian Splatting, even under sparse observations.}
  \label{fig:pipeline}
\end{figure*}

\noindent\textbf{System Overview.}~
Our system follows a \textit{Dual-Field} active reconstruction system that unifies semantic cues from a Vision-Language Model (VLM) with geometric cues from a feed-forward neural reconstructor \cite{vggt}. The semantic stream uses a VLM\cite{internvl3} to produce a complementary semantic uncertainty field. Both streams are fused on a shared voxel grid to produce a unified 3D uncertainty field, which guides the active viewpoint selection strategy. The entire pipeline is shown in Figure~\ref{fig:pipeline}.
 
Specifically, we present our system from three complementary components: 
a \textbf{Feed-forward Confidence Modeling} module for geometric uncertainty field estimation, a \textbf{Vision Language Model Understanding} module for high-level semantic uncertainty field estimation, and an \textbf{Active Viewpoint Selection} strategy for efficient view selection and scene exploration.

\subsection{Neural Feed Forward Confidence Modeling}
Neural feed-forward reconstruction, trained on large-scale datasets, exhibits outstanding reliability and robustness, even under sparse observations, while delivering rapid inference speed. By adopting a feed-forward 3D model, we introduce a \emph{geometric uncertainty field} derived from the 3D backbone~\cite{vggt}. Such a representation enables downstream modules to reason about reconstruction reliability and to guide active viewpoint selection effectively.

\vspace{0.3em}

\noindent\textbf{Feed Forward Neural 3D Perception Backbone.}~
We adopt \emph{VGGT}~\cite{vggt} as a transformer-based feed-forward geometry lifter that maps RGB tokens to dense pixel-level predictions in a single pass without online optimization. VGGT outputs a per-pixel depth confidence, which we interpret as predictive precision. We normalize this confidence into a [0,1] score and splat it onto a common voxel grid together with the corresponding depth. Pixel $\mathbf{x}=(x,y)$ with homogeneous coordinate $\tilde{\mathbf{x}}=[x,\,y,\,1]^{\top}$ is back-projected using intrinsics $K$ and camera pose $T_i\!\in\!\mathrm{SE}(3)$:
\[
\mathbf{X}_i(\mathbf{x}) \;=\; T_i\!\big(\,\hat D_i(\mathbf{x})\,K^{-1}\tilde{\mathbf{x}}\,\big),
\]
and the per-view uncertainty scores are aggregated across frames on the voxel grid to yield the geometric uncertainty field that drives selection.
\vspace{0.3em}

\noindent\textbf{Aleatoric Uncertainty Modeling.}~
The pretrained VGGT encoder is trained with a heteroscedastic objective that learns a per-pixel precision $c_i(\mathbf{x})$ to modulate the depth residual. In its simplified form\cite{uncertainties_bayesian},
\[
\mathcal{L}_{\text{depth}}
\;=\;
\sum_{\mathbf{x}}
\Big(
c_i(\mathbf{x})\,\ell_i(\mathbf{x})
\;-\;
\alpha\,\log c_i(\mathbf{x})
\Big),
\]
where $\ell_i(\mathbf{x})$ denotes the depth discrepancy. 
This objective encourages high confidence $c_i$ in reliable regions and lower values in ambiguous areas, 
allowing the pretrained confidence map to act as a natural proxy for aleatoric, input-dependent uncertainty during inference. 

We apply a simple monotonic normalization and 3D back-projection to integrate this uncertainty for fusion and view selection. 
Leveraging this uncertainty-aware feed-forward design, our VGGT backbone provides robust 3D perception from sparse observations while maintaining efficient inference for active view planning.

\subsection{Vision Language Model for High-Level Reasoning}
\label{sec:vlm}
Geometric confidence from the vision backbone captures areas of high reconstruction ambiguity, but purely geometry-driven signals may miss semantically important regions that are difficult to reconstruct. To provide this complementary guidance, we induce a \emph{semantic uncertainty field} that leverages high-level cues from a Vision Language Model\cite{internvl3} and modulates them with vision-feature variability. The resulting per-image map is then lifted into 3D and fused across views to guide efficient viewpoint selection.
\vspace{0.3em}

\noindent\textbf{Uncertainty Region Analysis.}~
\label{sec:uncertainty_analysis}
We leverage the Vision–Language Model as a semantic prior to identify regions where additional observations are likely to improve reconstruction quality, such as occluded areas, thin structures, reflective or textureless surfaces.
To obtain spatially grounded and consistent predictions, we design a structured prompt that divides the image into fixed coarse grids and asks the VLM to output a small number of region tuples, each associated with a \emph{category}, including OCCLUSION, GEOMETRIC, LIGHTING, BOUNDARY, TEXTURE and a corresponding \emph{priority} score.
This constrained format reduces the free-form variability of large language models and allows the responses to be mapped deterministically to image coordinates.

\noindent\textbf{Parsing Uncertainty Regions.}~
\label{sec:parse_uncertainty}
Each VLM-predicted region is converted into a soft spatial mask $M_k(u)\!\in\![0,1]$ with Gaussian edge tapering and mild dilation for recall. 
We assign each region a calibrated weight based on its category and priority:
\[
W_i(u) \;=\; \sum_{k=1}^{K} \alpha_{\mathrm{type}_k}\,\beta_{\mathrm{prio}_k}\,M_k(u),
\]
where $\alpha$ and $\beta$ are fixed coefficients for region type and priority, respectively. 
The aggregated map $W_i(u)$ is normalized per image to $[0,1]$ for stability.

Let $\sigma_i(u)$ denote the feature-level uncertainty from the vision backbone.
The final semantic-modulated uncertainty is
\[
U^{\mathrm{sem}}_i(u)
\;=\;
\mathrm{Norm}\!\big(\sigma_i(u)\,[1+\lambda\,W_i(u)]\big),
\]
where $\lambda$ controls modulation strength. 
The resulting $U^{\mathrm{sem}}_i(u)$ serves as a dense semantic field that integrates with geometric confidence in 3D to guide viewpoint selection. By projecting this semantic field into 3D and fusing it with the feed-forward geometric confidence, our system obtains a unified uncertainty map that guides the agent toward the most informative viewpoints.

Meanwhile, to ensure that the reconstruction process does not remain confined to the initially observed views, we assign a global uncertainty weight to all voxels.

\vspace{0.3em}
\subsection{Active Viewpoint Selection Strategy }
\noindent\textbf{Problem Formulation.}~
With scores from the dual fields, we could form the active viewpoint selection problem as an information gain problem. We characterize selection as maximizing the expected reduction in the fused uncertainty field within the cone of vision of a candidate pose. 

\vspace{0.3em}
\noindent\textbf{Visibility Gate.}~
Visibility is a pose-conditioned operator shared by the semantic and geometric fields. We remove out-of-frustum voxels with a deterministic frustum test. To capture occlusions, we precompute a probabilistic FOV mask at each seed over a coarse yaw/pitch grid. 
The mask is generated using Monte Carlo ray sampling with first-hit termination within the cone of vision, following the approach of \cite{worldmem}. During selection we reuse these cached masks instead of reshooting rays. The mask gates all scoring and fusion so that utility is assigned only to potentially observable content.

\vspace{0.3em}
\noindent\textbf{Frustum-based Uncertainty Decay.}~
\label{sec:frustum}
After committing a view, we multiplicatively reduce the fused uncertainty within the corresponding precomputed frustum mask. This couples selection with evidence accumulation, making already explained regions less attractive, while residual visibility-gated uncertainty guides the policy toward novel surfaces. Seeds are re-evaluated under the decayed field, and the same location is revisited only when alternative orientations remain informative.

\vspace{0.3em}
\noindent\textbf{Viewpoint Candidates Generation.}~
Selection is split into \emph{center} and \emph{direction} stages. 
We first voxelize the workspace and treat valid voxel centers as candidate camera seeds. For each seed, visibility masks over a small set of view orientations are precomputed via Monte Carlo ray sampling and cached for efficient reuse. Seeds are first pre-processed to be maintained in a max-priority queue, which enables efficient greedy selection of the most informative candidate at each iteration. The selection proceeds iteratively in a greedy fashion: at each iteration, the top-ranked seed is popped, a compact fan of candidate orientations and ranges is instantiated, and each pose is scored by fast accumulation over the cached visibility masks. Uncertainty within the selected view's frustum is decayed by a constant factor, and affected seeds are re-keyed and re-queued. This greedy loop continues until the view budget is exhausted. The full procedure is summarized in Algorithm~\ref{alg:avs}.

\begin{algorithm}[t]
\small
\caption{\textbf{Active View Selection with Dual Fields Uncertainty Guidance}}
\label{alg:avs}
\SetAlgoNoLine
\SetKwInOut{Input}{Input}\SetKwInOut{Output}{Output}
\Input{Workspace $\mathcal{W}$, observations $\mathcal{O}$, view budget $N$, MC rays $R_{\mathrm{pre}}$}
\Output{Next viewpoints $\mathcal{V}_{\mathrm{next}}$}

\BlankLine
\textbf{Precompute:}
\begin{enumerate}\itemsep2pt
  \item \textbf{Voxelize} $\mathcal{W}$ for candidate seeds.
  \item Build \textbf{dual fields} (geometry $\mathcal{F}_g$, semantics $\mathcal{F}_s$); fuse to utility map $\mathcal{U}$.
  \item Run \textbf{Monte Carlo} rays ($R_{\mathrm{pre}}$) per seed and orientation bin to estimate visibility; cache \textbf{FOV masks}.
  \item Initialize a \textbf{priority queue} $\mathcal{Q}$ using mask–weighted utility bounds (with distance prior).
\end{enumerate}

\BlankLine
\textbf{Iterate:}
\For{$t = 1$ \KwTo $N$}{
  Pop top seed $s^\star$ from $\mathcal{Q}$\;
  Evaluate poses at $s^\star$ via \textbf{mask-only scoring} on cached masks\;
  Commit best pose to $\mathcal{V}_{\mathrm{next}}$\;
  Apply \textbf{frustum-based decay} to $\mathcal{U}$\;
  Update priority queue for local neighbors under decayed $\mathcal{U}$; reinsert if promising (with light NMS)\;
}
\Return{$\mathcal{V}_{\mathrm{next}}$}
\end{algorithm}
\FloatBarrier
\begin{table*}[t]
\centering
\caption{Scene-level results on the Replica dataset. We report PSNR$\uparrow$, SSIM$\uparrow$, and LPIPS$\downarrow$. Our method is bolded.}
\label{tab:scene_level}
\vspace{0.6em}
\setlength{\tabcolsep}{4pt}
\renewcommand{\arraystretch}{1.15}
\adjustbox{max width=\textwidth}{
\begin{tabular}{lccc|ccc|ccc|ccc}
\toprule
\multirow{2}{*}{\textbf{Method}} &
\multicolumn{3}{c|}{\textbf{room0}} &
\multicolumn{3}{c|}{\textbf{office0}} &
\multicolumn{3}{c|}{\textbf{office2}} &
\multicolumn{3}{c}{\textbf{office4}} \\
\cmidrule(lr){2-4} \cmidrule(lr){5-7} \cmidrule(lr){8-10} \cmidrule(lr){11-13}
 & PSNR$\uparrow$ & SSIM$\uparrow$ & LPIPS$\downarrow$
 & PSNR$\uparrow$ & SSIM$\uparrow$ & LPIPS$\downarrow$
 & PSNR$\uparrow$ & SSIM$\uparrow$ & LPIPS$\downarrow$
 & PSNR$\uparrow$ & SSIM$\uparrow$ & LPIPS$\downarrow$ \\
\midrule
Random      & 28.17 & 0.821 & 0.152 & 32.35 & 0.826 & 0.152 & 27.75 & 0.837 & 0.157 & 26.19 & 0.829 & 0.166 \\
VLM-based   & 27.21 & 0.808 & 0.193 & 24.92 & 0.791 & 0.201 & 25.91 & 0.810 & 0.187 & 23.90 & 0.802 & 0.190 \\
FisherRF    & 29.11 & 0.832 & 0.151 & 27.13 & 0.825 & 0.156 & 28.20 & 0.840 & 0.142 & 27.79 & 0.827 & 0.152 \\
\textbf{Ours w/o VLM} & 28.53 & 0.855 & 0.123 &
31.75 & 0.842 & 0.134 &
28.00 & 0.853& 0.128&
30.06 & 0.847 & 0.132 \\
\textbf{Ours} & \textbf{29.23} & \textbf{0.867} & \textbf{0.110} &
\textbf{32.98} & \textbf{0.855} & \textbf{0.120} &
\textbf{28.70} & \textbf{0.862} & \textbf{0.115} &
\textbf{31.79} & \textbf{0.858} & \textbf{0.118} \\
\bottomrule
\end{tabular}
}
\vspace{-0.5em}
\end{table*}

\begin{table*}[htbp!]
\centering
\caption{
Object-level results under different scene complexities.
We report PSNR$\uparrow$, SSIM$\uparrow$, and LPIPS$\downarrow$.
Our method is boled.
}
\label{tab:object_level}
\setlength{\tabcolsep}{3.0pt} 
\renewcommand{\arraystretch}{1.05} 
\begin{tabular}{lccc|ccc|ccc|ccc}
\toprule
\multirow{2}{*}{Method} &
\multicolumn{3}{c|}{Single-object} &
\multicolumn{3}{c|}{5-objects} &
\multicolumn{3}{c|}{7-objects 1} &
\multicolumn{3}{c}{7-objects 2} \\
\cmidrule(lr){2-4} \cmidrule(lr){5-7} \cmidrule(lr){8-10} \cmidrule(lr){11-13}
 & PSNR$\uparrow$ & SSIM$\uparrow$ & LPIPS$\downarrow$ 
 & PSNR$\uparrow$ & SSIM$\uparrow$ & LPIPS$\downarrow$ 
 & PSNR$\uparrow$ & SSIM$\uparrow$ & LPIPS$\downarrow$
 & PSNR$\uparrow$ & SSIM$\uparrow$ & LPIPS$\downarrow$ \\

\midrule
Random & 31.37 & 0.870 & 0.088 & 29.66 & 0.882 & 0.105 & 29.61 & 0.853 & 0.143 & 32.24 & 0.892 & 0.099 \\
Uniform & 32.15 & 0.880 & 0.088 & 30.69 & 0.884 & 0.109 & 29.86 & 0.861 & 0.146 & 32.25 & 0.894 & 0.099 \\
VLM-based & 26.29 & 0.859 & 0.125 & 21.80 & 0.759 & 0.218 & 22.44 & 0.769 & 0.191 & 24.93 & 0.819 & 0.169 \\
Air-Embodied & 30.35 & 0.885 & 0.102 & 30.35 & 0.887& 0.101& 28.35 & 0.823 & 0.197 & 29.85 & 0.795&0.121 \\
\textbf{Ours w/o VLM} & 29.74 & 0.873 & 0.096 & 31.43 & 0.890 & 0.101 & 32.39 & 0.891 & 0.083 & 31.48 & 0.880 & 0.107 \\
\textbf{Ours} & \textbf{31.59} & \textbf{0.893} & \textbf{0.093} & \textbf{31.86} & \textbf{0.893} & \textbf{0.098} & \textbf{33.44} & \textbf{0.899} & \textbf{0.081} & \textbf{32.69} & \textbf{0.902} & \textbf{0.085} \\
\bottomrule
\end{tabular}
\end{table*}

Generally, We couple dual-field utility with precomputed visibility. Voxelization bounds the search, and cached frustum masks with a priority queue focus scoring on observable high-gain seeds. Monte Carlo is used for mask construction or occasional refresh. Frustum-based uncertainty decay turns evidence into reduced utility, yielding a budget-aware policy that balances exploration and exploitation. These design choices enable efficient, robust view selection even under sparse visual observations.

\begin{figure*}[t]
  \centering
  \includegraphics[width=\linewidth]{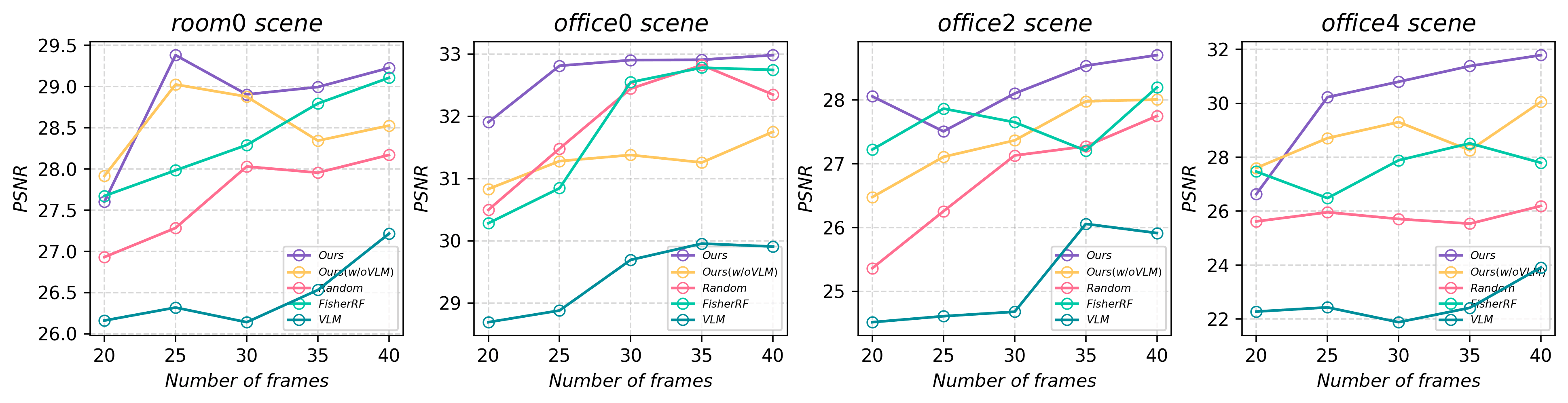}
  \caption{PSNR as the number of input frames increases under different view-selection policies in the scene-level setting.}
  \label{fig:efficiency_scene}
\end{figure*}

\begin{figure*}[htbp!]
  \centering
  \includegraphics[width=\linewidth]{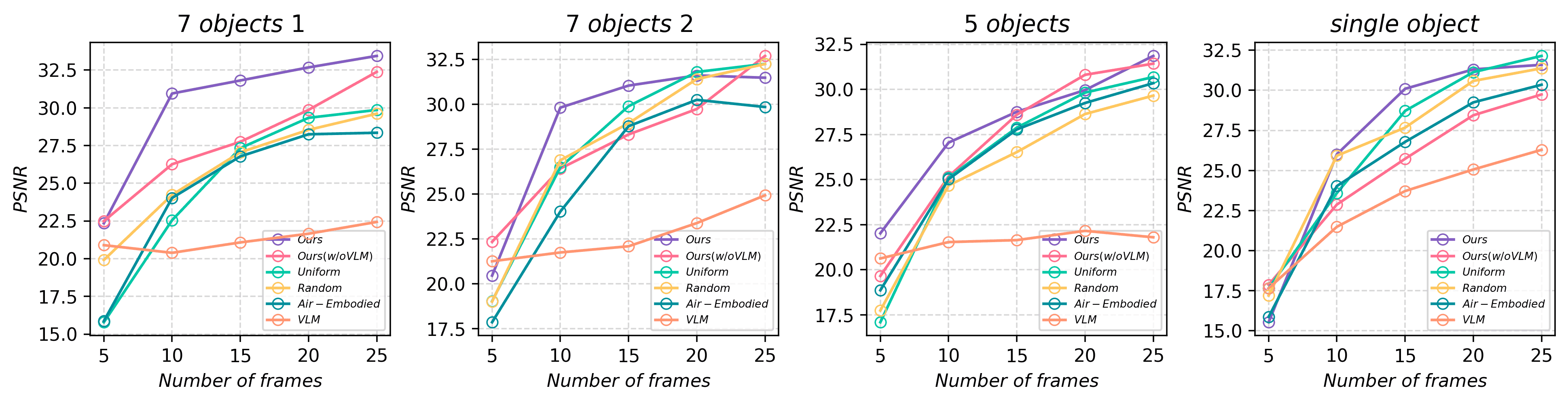}
  \caption{PSNR as the number of input frames increases under different view-selection policies in the object-level setting.}
  \label{fig:efficiency_object}
\end{figure*}

\section{Experiments}
\label{sec:formatting}
\subsection{Data Generation}
\label{sec:data_generation}
We construct a unified active reconstruction benchmark covering both object-centric and scene-level regimes. We conduct experiments at both object-level and scene-level within simulated environments, using CoppeliaSim and Habitat \cite{habitat1,habitat2,habitat3} simulators, respectively.
\vspace{0.5em}

\noindent\textbf{Scene-level.}~ 
For scene-level experiments, we focus on single-room indoor environments. We adopt the Replica dataset~\cite{replica} as our testing scenario and follow the data generation protocol of Semantic-NeRF~\cite{semanticnerf}. Specifically, we generate the dataset by replaying the camera trajectories and split the rendered frames into an initial observation dataset and test sets for fair evaluation.
\vspace{0.5em}

\noindent\textbf{Object-level.}~ 
For object-level experiments, we focus on single and multi-object tabletop environments. We adopt the OmniObject3D dataset~\cite{omniobject3d}. We craft single-object and multi-object scenarios specifically. The initial observation dataset and test set are uniformly sampled on a hemisphere centered on the object. Due to the increased complexity of the 7-object case, we create two distinct scenarios for evaluation.



\subsection{Experiment Setup}
We evaluate active 3D reconstruction under a fixed view budget. 
Let $\mathcal{O}_0$ denote the initial observations and $T$ the total view budget. At each episode, the agent can acquire up to $T-|\mathcal{O}_0|$ additional views. We consider two settings: object-level with $|\mathcal{O}_0| = 4$ and $T = 25$, and scene-level with $|\mathcal{O}_0| = 15$ and $T = 40$.
Reconstruction quality is evaluated on the resulting scene estimate $\hat{\mathcal{G}}(\mathcal{S})$ using 3D Gaussian Splatting; specifically, we adopt PGSR~\cite{pgsr} as the downstream representation.

\subsection{Metrics}
We quantitatively evaluate the reconstruction quality of 3D Gaussian Splatting using PSNR, SSIM \cite{ssim}, and LPIPS \cite{lpips}, which jointly measure pixel-level accuracy, structural consistency, and perceptual similarity. These metrics are widely adopted in image reconstruction and novel view synthesis tasks, and have been proven to reliably reflect perceptual and structural fidelity.
\subsection{Baselines}
\label{subsec:baselines}

We compare our proposed method against a set of representative active reconstruction strategies. Several baselines are common to both experimental levels, while others are specialized for either the object-level or scene-level task, as detailed below.

\begin{itemize}[wide, labelindent=0pt]
    \item \textbf{Random Sampling:} A non-active baseline where a fixed number of views are collected by sampling poses randomly.
    

    \item \textbf{Uniform Sampling:} A non-active baseline that samples poses uniformly. 
    

    \item \textbf{VLM-based (Naive):} An active baseline employing a VLM. (1) Reasoning: The VLM is prompted with the current view. (2) Planning: An LLM interprets the VLM's output and issues a simple, non-metric movement command.
\end{itemize}

\begin{itemize}[wide, labelindent=0pt]
    \item \textbf{AIR-Embodied (Object-Level):} Air-Embodied is a 3D Gaussian-based active reconstruction framework. We implement a version of the AIR-Embodied framework \cite{air}. To isolate its planning strategy from its interaction capabilities, we disable the object manipulation module. 
    

    \item \textbf{FisherRF (Scene-Level):} 
    FisherRF\cite{fisherrf} selects exploration paths by maximizing the Expected Information Gain, which is approximated from the Fisher Information of the 3DGS parameters.
\end{itemize}

\begin{figure*}[htbp!]
  \centering

  \includegraphics[width=\linewidth]{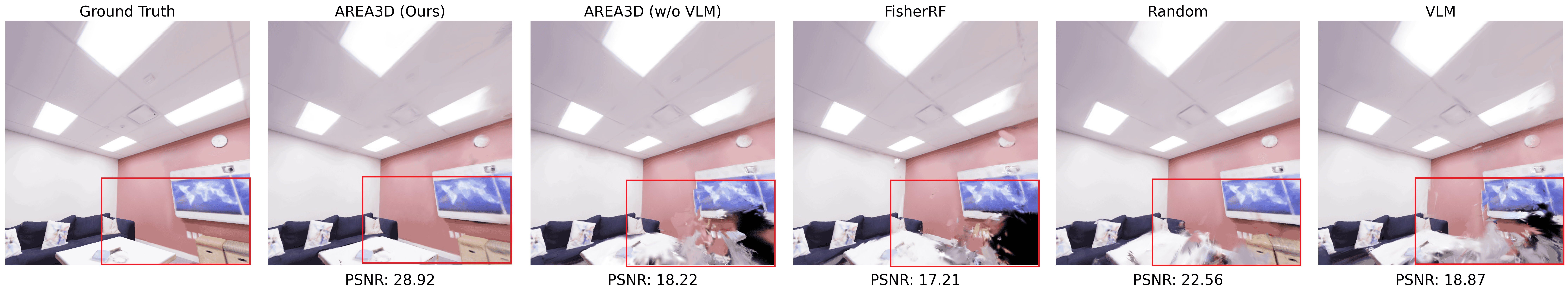}
  
  \includegraphics[width=\linewidth]{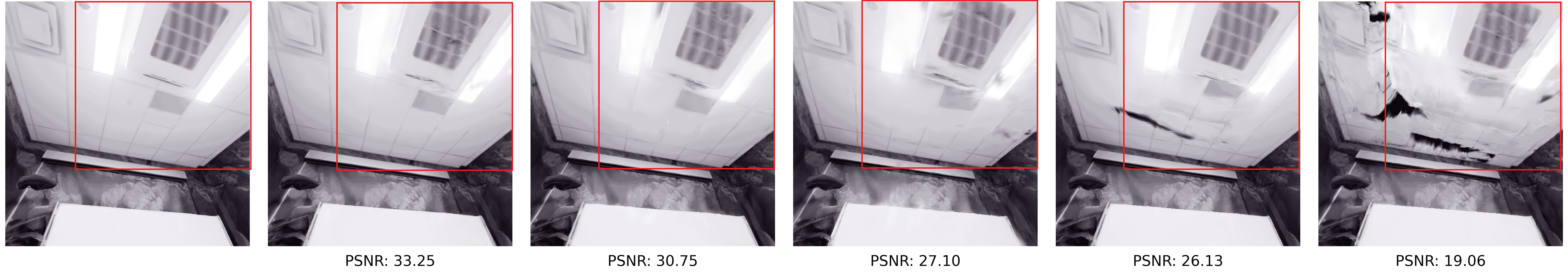}
  
  \includegraphics[width=\linewidth]{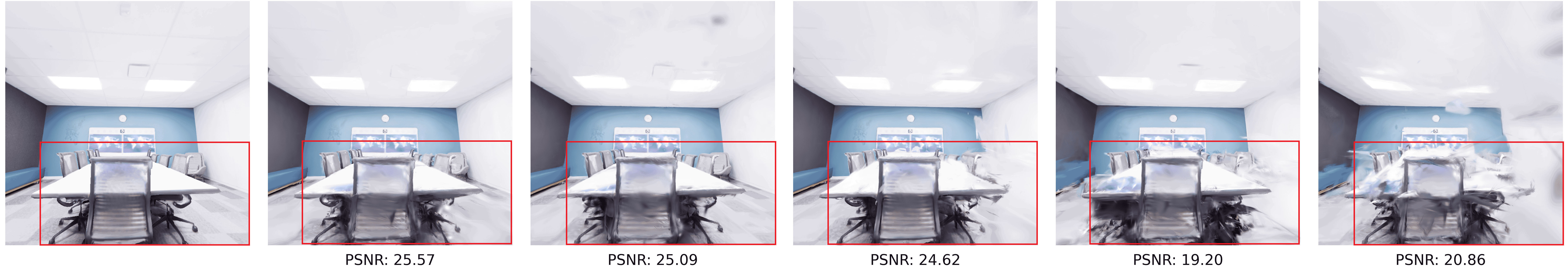}

  \vspace{-2mm}
  \caption{
    \textbf{Novel view synthesis results with different policies.}
  }
  \label{fig:NVS}
  \vspace{-3mm}
\end{figure*}

\subsection{Results}
\noindent\textbf{Cross-scale Experiment.}~ 
To evaluate the our model's generalizability across different spatial scales, we conduct cross-scale experiments in both single-room indoor and tabletop scenarios. This setting examines whether our framework adapts to variations in scene scale, geometry complexity, and object distribution. In the tabletop setup, we include single-object and multi-object configurations to further test robustness under different levels of scene complexity. Results are summarized in Tables~\ref{tab:scene_level} and \ref{tab:object_level}.

Figure~\ref{fig:NVS} provides visualizations of novel view synthesis for different policies, highlighting that our approach selects viewpoints that effectively capture regions left unobserved by other strategies.

\noindent\textbf{Effectiveness Compared with Baselines.}~
To verify the effectiveness of our method, we compare it with representative baselines under identical experimental setups. 
We additionally plot PSNR as a function of the number of acquired viewpoints to illustrate the efficiency of each policy in achieving high-quality reconstruction under a limited view budget. As shown in Figures~\ref{fig:efficiency_scene} and~\ref{fig:efficiency_object}, our method generally achieves higher PSNR with a comparable or smaller number of viewpoints, for example, 10 frames in the object-level experiments and 25 frames in the scene-level experiments, demonstrating efficient reconstruction.
Notably, the superiority of our policy is more evident in multi-object scenarios, suggesting that it more effectively captures complex geometries, which is a capability enabled by the synergy of 3D feed-forward perception and semantic-level guidance.

\noindent\textbf{Scalability.}~
We report the wall-clock runtime of AREA3D under different view budgets and voxel resolutions. As shown in Table~\ref{tab:eff_scaling}, the total runtime increases only slightly as the view budget $T$ grows, indicating that our method scales well with the number of input views. In contrast, decreasing the voxel size $v$ leads to increased runtime due to the higher computational cost associated with finer spatial discretization. Nevertheless, the overall runtime remains within a reasonable range, demonstrating the practical efficiency and scalability of our approach.
\begin{table}[t]
\centering
\caption{Ablation of Feed-Forward Perception and VLM Guidance at object and scene levels. Our method is bolded.}
\label{tab:ablation}
\setlength{\tabcolsep}{8pt}
\renewcommand{\arraystretch}{0.85}
\begin{tabular}{ccccc}
\toprule
\multicolumn{2}{c}{\textbf{Components}} & \multicolumn{3}{c}{\textbf{Performance}} \\
\cmidrule(lr){1-2} \cmidrule(lr){3-5}
Feed-Forward & VLM & PSNR$\uparrow$ & SSIM$\uparrow$ & LPIPS$\downarrow$ \\
\midrule
\multicolumn{5}{l}{\textit{Object-level}} \\
\ding{55} & \ding{51} & 29.02 & 0.844 & 0.202  \\
\ding{51} & \ding{55} & 31.56 & 0.896 & 0.091   \\
\ding{51} & \ding{51} & \textbf{32.09} & \textbf{0.886} & \textbf{0.102} \\
\midrule
\multicolumn{5}{l}{\textit{Scene-level}} \\
\ding{55} & \ding{51} & 29.10 & 0.839 & 0.115 \\
\ding{51} & \ding{55} & 31.26 & 0.884 & 0.097 \\
\ding{51} & \ding{51} & \textbf{32.40} & \textbf{0.897} & \textbf{0.089} \\
\bottomrule
\end{tabular}
\end{table}

\subsection{Ablation}
We conduct ablations on the two main components of our framework: Feed-Forward Perception Field and Vision-Language Guidance Field. To evaluate their individual contributions, we disable each component by setting its corresponding weight to zero while keeping the rest of the pipeline unchanged. The results consistently validate the effectiveness of our framework design. The results are summarized in Table~\ref{tab:ablation}.

\begin{table}[t]
\centering
\caption{\textbf{Scalability.} Wall-clock time.}
\label{tab:eff_scaling}
\small
\renewcommand{\arraystretch}{0.95}
\vspace{-0.3cm}

\textit{(a) Scaling with view budget $T$ (fixed voxel size $v=0.5$)}
\vspace{0.3em}

{\setlength{\tabcolsep}{4pt}
\begin{tabular}{cc cc cc cc}
\toprule
$T$ & Total(s)$\downarrow$ & $T$ & Total(s)$\downarrow$ & $T$ & Total(s)$\downarrow$ & $T$ & Total(s)$\downarrow$ \\
\midrule
5  & 134.9   & 10 & 135.1   & 15 & 135.5   & 20 & 135.8 \\
25 & 137.6 & 30 & 137.9 & 35 & 138.1 & 40 & 139.2 \\
\bottomrule
\end{tabular}}

\vspace{0.5em}
\textit{(b) Scaling with voxel size $v$ at fixed $F=25$}
\vspace{0.3em}

\begin{tabular*}{\columnwidth}{@{\extracolsep{\fill}} lcccc}
\toprule
 & $v=0.5$ & $v=0.4$ & $v=0.3$ & $v=0.2$ \\
\midrule
Total(s)$\downarrow$      & 138 & 139 & 159 & 207 \\
\bottomrule
\end{tabular*}
\end{table}


\section{Conclusion}
In this paper, we present \textbf{AREA3D}, an active 3D reconstruction agent that integrates a feed-forward 3D perception model with vision-language guidance. 
By leveraging the complementary strengths of pretrained feed-forward 3D models and vision-language models, AREA3D enables intelligent and efficient active view selection without requiring optimization-based policies. 
Extensive experiments demonstrate that our framework achieves state-of-the-art performance in novel view synthesis in sparse views using 3D Gaussian reconstruction, significantly reducing the number of required views while maintaining high reconstruction quality. 

\section{Acknowledgement}
This work was done while Tianling Xu was an intern at Harvard University. The work was partially supported by NIH grant R01HD104969 and NSF grant CRCNS-2309041.

\clearpage
{
    \small
    \bibliographystyle{ieeenat_fullname}
    \bibliography{main}
}

\newpage
\setcounter{page}{1}
\maketitlesupplementary
\section{Dataset and Benchmark}
\label{sec:Dataset}
\begin{figure*}[!t]
  \centering
  \includegraphics[width=\linewidth]{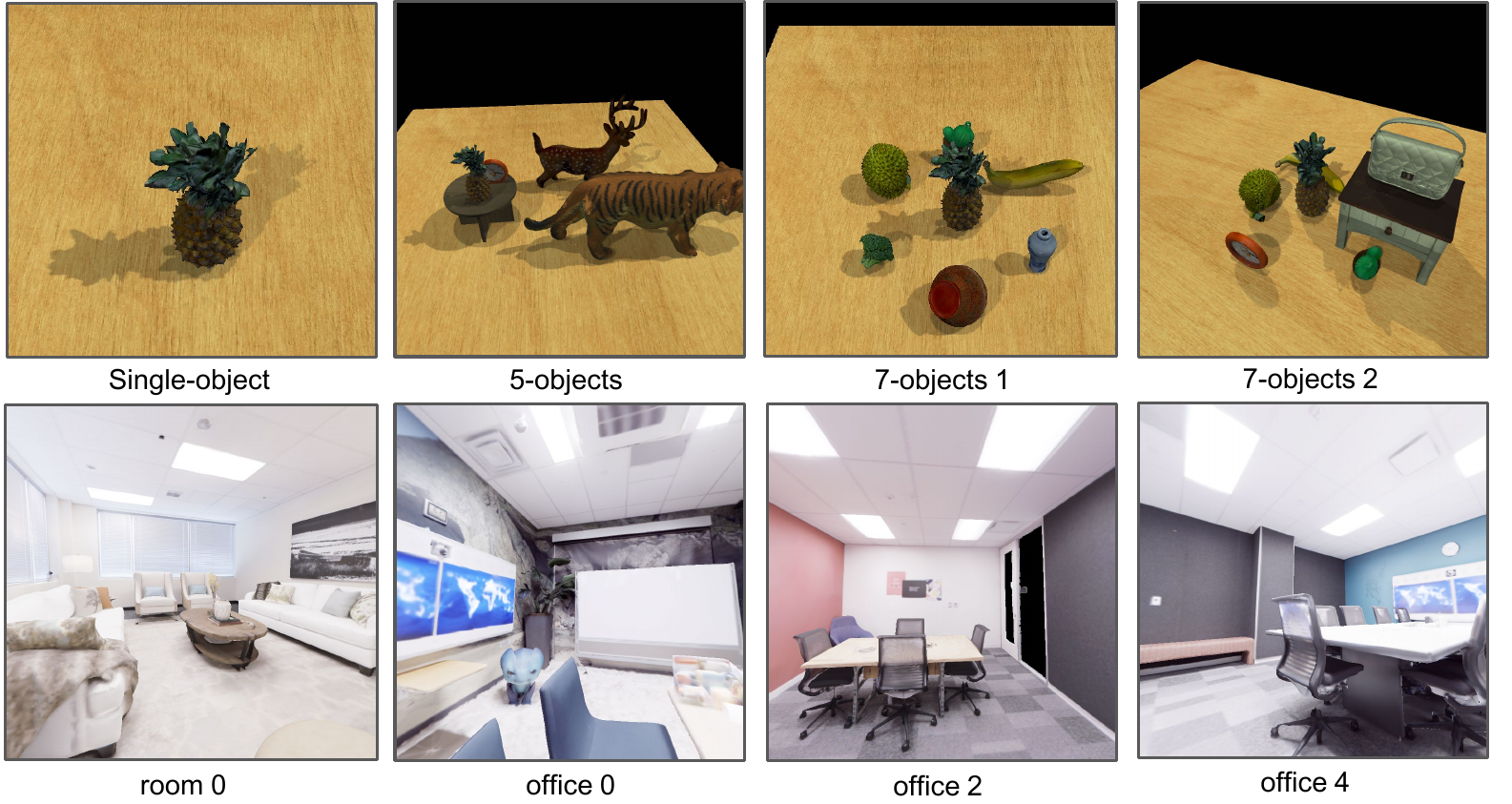}
  \vspace{-23pt} 
  \caption{Four single-room scenes that capture diverse indoor layouts, and four tabletop scenes featuring object-centric setups with rich geometric details and occlusions}
  \label{fig:dataset}
\end{figure*}

In Sec.~\ref{sec:data_generation} of the main paper, we introduce our unified benchmark for active 3D reconstruction. Here we provide the complete details of the dataset configuration and scene construction. As illustrated in Fig.~\ref{fig:dataset}, we include eight scenes in total: four single-room scenes that capture diverse indoor layouts, and four tabletop scenes featuring object-centric setups with rich geometric details and occlusions. These scenes are used consistently across baselines and ablations, enabling fair comparison under the same camera budget.

\section{Implementation Details}

\noindent\textbf{Systematic Prompt for VLM.}~
In Sec.~\ref{sec:vlm} we describe how the VLM output is fused with geometric uncertainty.
Here we detail the concrete prompt used in practice.
At the beginning of each episode, the agent collects $\mathcal{O}_0$ initial RGB views, and we query the VLM once with all $\mathcal{O}_0$ frames.
For each image, the field of view is divided into a coarse grid (horizontal: left, center-left, center-right, right; vertical: top, middle, bottom), and the VLM is asked to return a ranked list of regions, each with a location, an uncertainty type (\texttt{OCCLUSION}, \texttt{GEOMETRIC}, \texttt{LIGHTING}, \texttt{BOUNDARY}, or \texttt{TEXTURE}), a priority level (\texttt{HIGH}, \texttt{MEDIUM}, \texttt{LOW}), and a short natural-language justification.
The textual output is then parsed into per-pixel importance maps and lifted into a 3D, visibility-aware uncertainty field via 2D-to-3D unprojection.

For completeness, we show the instruction given to the VLM.
We provide all $\mathcal{O}_0$ initial images together with the following text:

\begin{quote}\small
You are an expert visual analyzer for active 3D reconstruction.
You will be given several RGB images (initial observations) from the same scene.
For each image, independently identify regions that require additional viewpoints for complete 3D reconstruction.

\medskip
\noindent\textbf{Coordinate system.}
Divide each image into a $4 \times 3$ grid:
horizontal positions are \emph{left}, \emph{center-left}, \emph{center-right}, and \emph{right};
vertical positions are \emph{top}, \emph{middle}, and \emph{bottom}.
Example locations include ``left-top'', ``center-left-middle'', ``right-bottom'', and ``center-right-top''.

\medskip
\noindent\textbf{Uncertainty categories (ranked by priority).}
\begin{itemize}[leftmargin=*]
\item \textbf{OCCLUSION (high):} hidden or blocked surfaces; regions behind furniture, walls, or large objects; back faces only visible from narrow viewing angles.
\item \textbf{GEOMETRIC (high):} thin structures, surfaces at grazing angles, complex curved shapes, reflective or transparent materials.
\item \textbf{LIGHTING (medium):} deep shadows, overexposed areas, strong highlights, blur, very low-contrast regions.
\item \textbf{BOUNDARY (medium):} objects cut by image borders, incomplete views, extreme tangential angles.
\item \textbf{TEXTURE (low):} textureless, repetitive, or very low-contrast regions.
\end{itemize}

\medskip
\noindent\textbf{Output format.}
For each image, list 5--8 regions in decreasing order of importance.
Each region should be summarized in one line with the following fields:

\begin{itemize}[leftmargin=*]
\item \texttt{REGION:} location using the grid notation (e.g., ``center-left-middle'').
\item \texttt{TYPE:} one of \texttt{OCCLUSION}, \texttt{GEOMETRIC}, \texttt{LIGHTING}, \texttt{BOUNDARY}, \texttt{TEXTURE}.
\item \texttt{PRIORITY:} \texttt{HIGH} (must observe), \texttt{MEDIUM} (should observe), or \texttt{LOW} (nice to observe).
\item \texttt{SIZE:} \texttt{small} ($<10\%$), \texttt{medium} ($10$--$25\%$), or \texttt{large} ($>25\%$) of the image.
\item \texttt{REASON:} 1--2 sentences explaining why extra viewpoints are needed and what 3D information is currently missing.
\end{itemize}
\end{quote}

\noindent\textbf{Parsing Uncertainty Regions.}~
In Sec.~\ref{sec:parse_uncertainty} of the main paper we define the 2D spatial weight map
\begin{equation}
W_i(u)
=
\sum_{k} \alpha_{\mathrm{type}_k}\,\beta_{\mathrm{prio}_k}\,M_k(u),
\end{equation}
and the semantic modulation;
\begin{equation}
U^{\mathrm{sem}}_i(u)
=
\mathrm{Norm}\big(\sigma_i(u)\,[1 + \lambda\,W_i(u)]\big).
\end{equation}
In our implementation, we employ a fixed weighting scheme that reflects the relative importance of different priorities and remains constant across all experiments.

We set the priority and size-dependent coefficients to fixed values
summarized in Table~\ref{tab:vlm_weights}; the same settings are used
for all experiments.
After aggregating over regions, $W_i(u)$ is normalized per image to
$[0,1]$.

\begin{table}[t]
\centering
\small
\caption{Coefficients for VLM region priority, size, and modulation.}
\label{tab:vlm_weights}
\begin{tabular}{lll}
\hline
Symbol & Meaning & Value \\
\hline
$\beta_{\text{HIGH}}$ & priority = HIGH   & $3.0$ \\
$\beta_{\text{MED}}$  & priority = MEDIUM & $1.5$ \\
$\beta_{\text{LOW}}$  & priority = LOW    & $0.5$ \\
\hline
$s_{\text{small}}$  & size = small   & $0.8$ \\
$s_{\text{medium}}$ & size = medium  & $1.0$ \\
$s_{\text{large}}$  & size = large   & $1.2$ \\
\hline
$\lambda$           & modulation strength & $1.0$ \\
\hline
\end{tabular}
\end{table}

\noindent\textbf{Frustum-based Uncertainty Decay.}~
In Sec.~\ref{sec:frustum} of the main paper we state that, after committing a view,
we multiplicatively reduce the fused uncertainty inside the
corresponding frustum.
Here we detail the decay rule used in our implementation.

Let $u_t(v)$ denote the fused 3D uncertainty at voxel center $v$ at step $t$.
Given a committed camera pose $T_w^c$, we first determine the set of
voxels whose centers fall inside the viewing frustum
$\mathrm{Frustum}(T_w^c)$, using the camera forward direction, a
field-of-view threshold, and a depth range consistent with view
rendering).
The uncertainty is then updated by
\begin{equation}
u_{t+1}(v) =
\begin{cases}
(1 - \eta)\,u_t(v), & v \in \mathrm{Frustum}(T_w^c),\\[2pt]
u_t(v),             & \text{otherwise},
\end{cases}
\end{equation}
i.e., all voxels inside the frustum are scaled by a constant decay
factor while others remain unchanged.

The hyperparameters used for this frustum-based decay are summarized
in Table~\ref{tab:frustum_decay_params} and are kept fixed for all
experiments.

\begin{table}[t]
\centering
\small
\caption{Hyperparameters for frustum-based uncertainty decay.}
\label{tab:frustum_decay_params}
\begin{tabular}{lll}
\hline
Symbol & Meaning            & Value \\
\hline
$\eta$           & decay factor          & $0.3$ \\
$\mathrm{FOV}$   & field of view         & $90^{\circ}$ \\
$\text{max\_depth}$ & maximum depth      & $5\,\mathrm{m}$ \\
\hline
\end{tabular}
\end{table}

\section{More Quantitative Results}
\vspace{0.3em}

\noindent\textbf{Overall Aggregate Performance.}~
To summarize performance on our benchmark, we aggregate the per-scene
PSNR, SSIM, and LPIPS reported in the main paper, separately for the
object-level and scene-level configurations. Averaged over all
object-level scenes, our policy attains 32.09 PSNR, 0.886 SSIM, and
0.102 LPIPS. On the scene-level benchmark, the corresponding averages
are 32.40 PSNR, 0.897 SSIM, and 0.089 LPIPS. These aggregated scores
provide a compact summary of our behavior on both parts of the benchmark
and are consistent with the per-scene comparisons in the main paper,
where our policy generally performs on par with or better than competing
methods under a fixed view budget.



\vspace{0.3em}

\noindent\textbf{Ablation on Global Initial Weight.}~
In Sec.~\ref{sec:parse_uncertainty} of the main paper we state that, to prevent the agent from
being confined to the initially observed views, we assign a global
initial uncertainty weight to all voxels.
Here we describe the exact form used in implementation and compare it
with a variant that removes this term.

Let $\hat{U}(v)$ denote the fused 3D uncertainty projected from the
2D semantic-modulated field.
Before view selection begins, each voxel is assigned a small additive
initial weight
\begin{equation}
\tilde{U}(v) = \hat{U}(v) + \gamma,
\end{equation}
where $\gamma$ is a constant offset that ensures non-zero uncertainty
for voxels not covered by the initial observation set.
This additive form is preserved throughout the reconstruction process and the global initial weight undergoes the same frustum-based decay as the other uncertainty components.
In practice, we use different values for the two benchmarks:
$\gamma = 0.01$ for the object-level setting and $\gamma = 0.005$ for
the scene-level setting.

We compare two configurations:
\textit{(i)} a baseline without global initial weight, where $\gamma = 0$; and
\textit{(ii)} our default setting with a non-zero initial weight
($\gamma = 0.01$ for object-level, $\gamma = 0.005$ for scene-level),
which preserves a minimal amount of residual uncertainty in unseen regions and encourages the policy to explore outside the initially observed frustum.

Quantitative results on both object-level and scene-level benchmarks are
reported in Table~\ref{tab:global_prior_ablation}.
Using a non-zero initial weight consistently improves viewpoint coverage
and leads to better long-range reconstruction quality.

\begin{table}[t]
\centering
\small
\caption{Ablation study of the global initial weight on both object-level and scene-level benchmarks.}
\label{tab:global_prior_ablation}
\setlength{\tabcolsep}{6pt}
\renewcommand{\arraystretch}{1.05}
\begin{tabular}{lccc}
\toprule
Setting & PSNR$\uparrow$ & SSIM$\uparrow$ & LPIPS$\downarrow$ \\
\midrule
\multicolumn{4}{l}{\textit{Object-level}} \\
$\gamma = 0$          & 29.212 & 0.859 & 0.120 \\
$\gamma = 0.01$ (ours) & \textbf{29.661} & \textbf{0.870} & \textbf{0.111} \\
\midrule
\multicolumn{4}{l}{\textit{Scene-level}} \\
$\gamma = 0$          &   27.845     &  0.837     & 0.153      \\
$\gamma = 0.005$ (ours) &   \textbf{28.265}     &  \textbf{0.848}     & \textbf{0.112}      \\
\bottomrule
\end{tabular}
\end{table}

\section{More Visualization Results}
\begin{figure*}[htbp!]
  \centering

  \includegraphics[width=\linewidth]{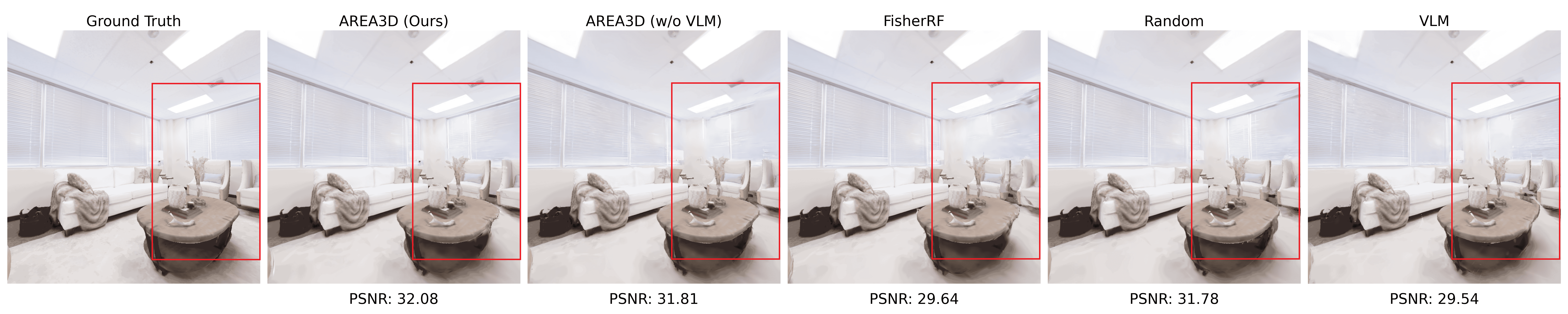}
  
  \includegraphics[width=\linewidth]{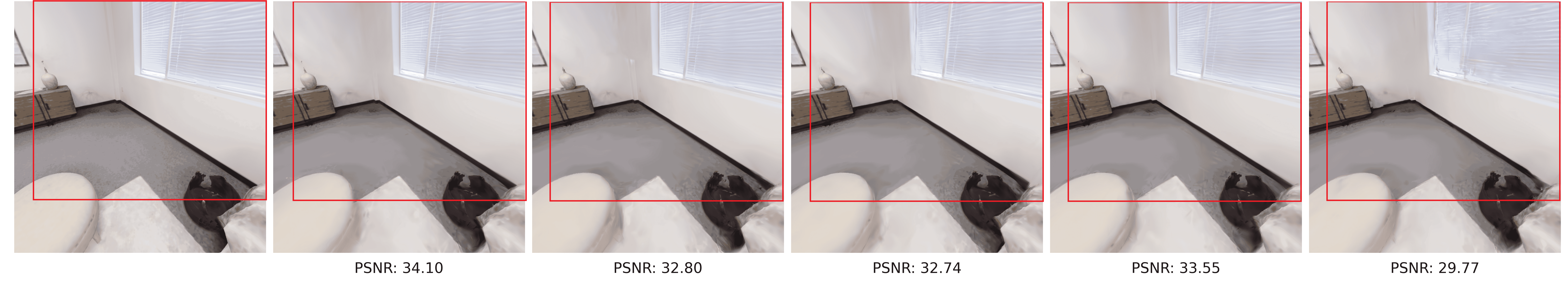}
  
  \includegraphics[width=\linewidth]{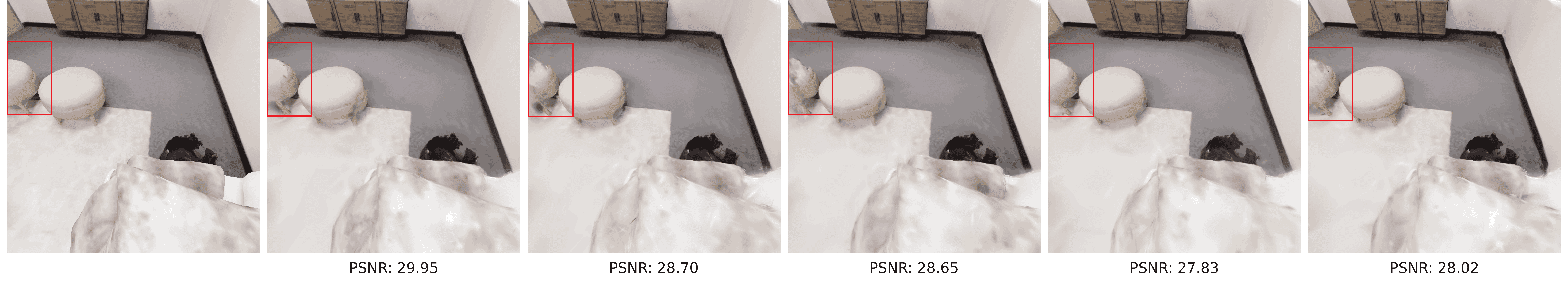}

  \includegraphics[width=\linewidth]{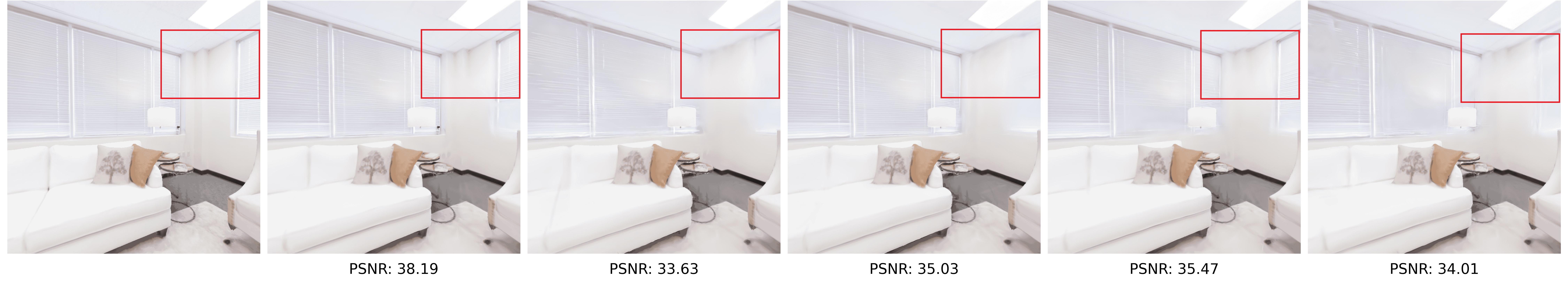}

  \includegraphics[width=\linewidth]{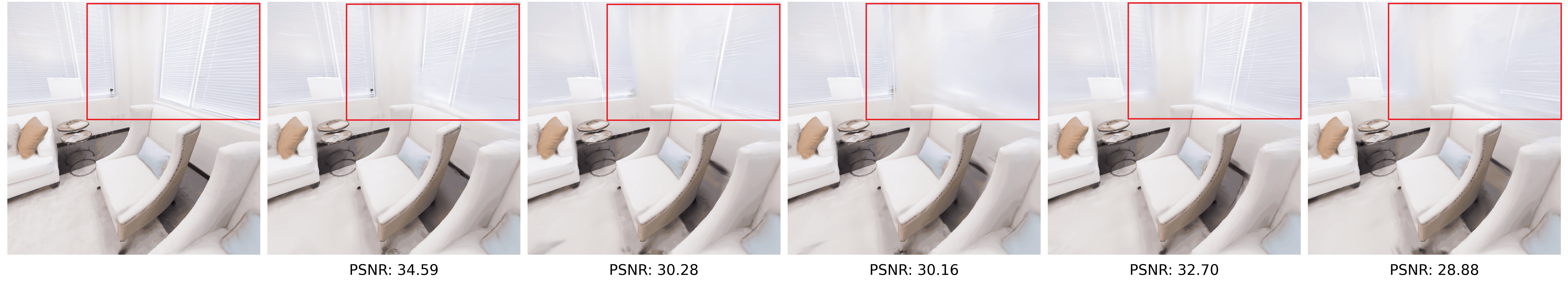}

  \includegraphics[width=\linewidth]{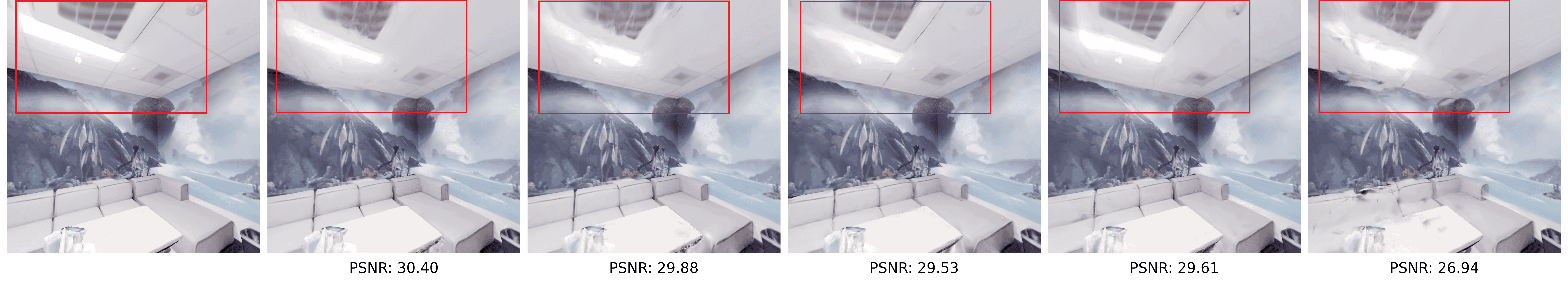}

  \vspace{-2mm}
  \caption{
    \textbf{Novel View Synthesis Results of different policies in scene-level.}
  }
  \label{fig:scene-NVS}
  \vspace{-3mm}
\end{figure*}

\begin{figure*}[t!]
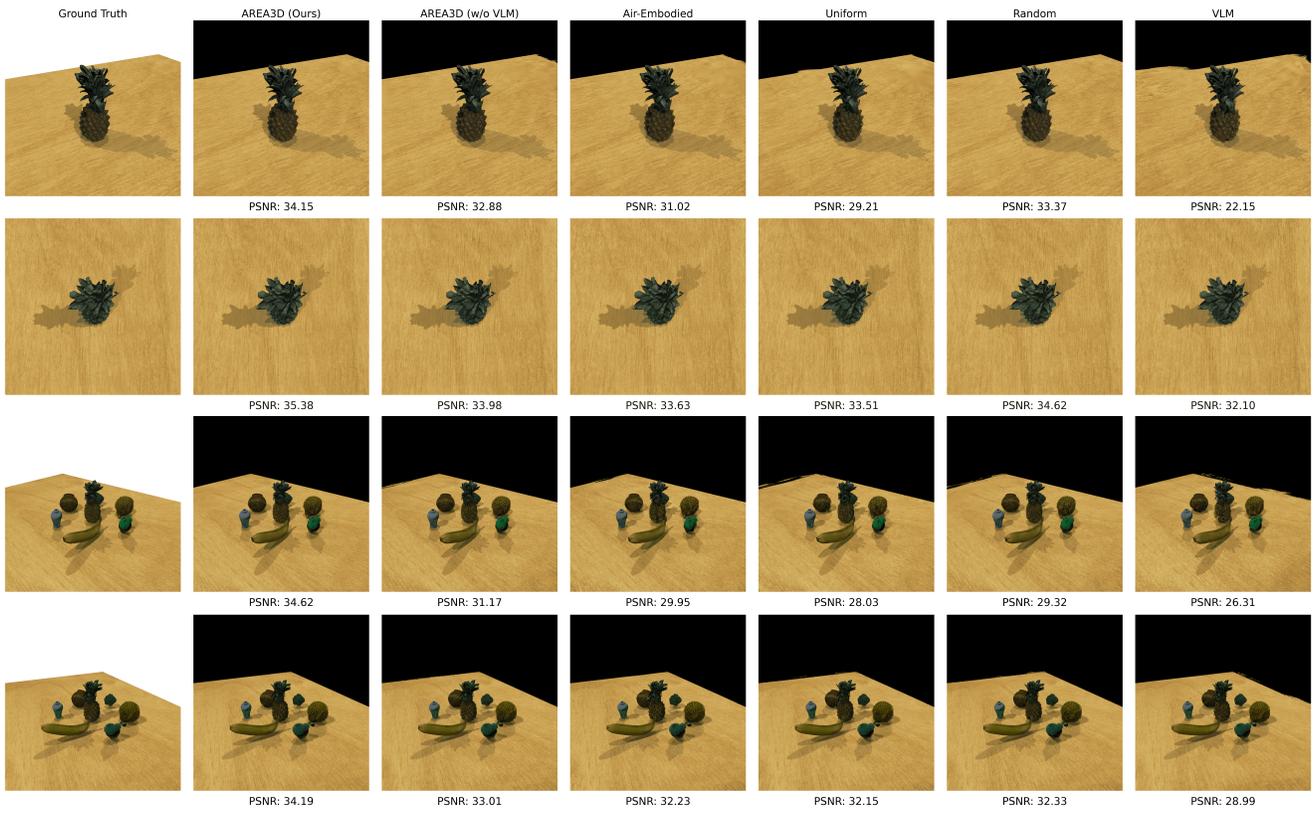

  \centering

  \includegraphics[width=\linewidth]{single_object.png}
  
  \includegraphics[width=\linewidth]{single_object2_crop.png}
  
  \includegraphics[width=\linewidth]{7_objects_crop.png}

  \includegraphics[width=\linewidth]{7_objects2_crop.png}

  \vspace{-2mm}
  \caption{
    \textbf{Novel View Synthesis Results of different policies in object-level.}
  }
  \label{fig:object-NVS}
  \vspace{-3mm}
\end{figure*}

Due to space constraints in the main paper, we only show three qualitative examples of novel view synthesis results obtained with 3D Gaussian Splatting under our active reconstruction policy. Here we provide additional visualizations covering both scene-level and object-level settings.
Each row compares our method with baselines on the same target view, as illustrated in Fig.~\ref{fig:scene-NVS} and Fig.~\ref{fig:object-NVS}.

\end{document}